\documentclass[utf8]{FrontiersinHarvard} % for articles in journals using the Harvard Referencing Style (Author-Date), for Frontiers Reference Styles by Journal: https://zendesk.frontiersin.org/hc/en-us/articles/360017860337-Frontiers-Reference-Styles-by-Journal
\usepackage{url,hyperref,lineno,microtype,subcaption}
\usepackage[onehalfspacing]{setspace}
\usepackage{tabularx}
\usepackage{multirow}
\usepackage{svg}
\usepackage{amsmath}
\usepackage{amssymb}
\usepackage[]{mdframed}

\def\keyFont{\fontsize{8}{11}\helveticabold }
\def\firstAuthorLast{Xu {et~al.}} %use et al only if is more than 1 author
\def\Authors{Mingle Xu\,$^{1,\dagger}$, Hyongsuk Kim\,$^{1,\dagger}$, Jucheng Yang\,$^{2,*}$, Alvaro Fuentes\,$^{1}$, Yao Meng\,$^{1}$, Sook Yoon\,$^{3,*}$, Taehyun Kim\,$^{4}$ and Dong Sun Park\,$^{1}$}
\def\Address{
$^{\dagger}$Equal Contribution. \\
$^{1}$Department of Electronic Engineering, Core Research Institute of Intelligent Robots, Jeonbuk National University, South Korea. \\
$^{2}$College of Artificial Intelligence, Tianjin University of Science and Technology, China. \\
$^{3}$Department of Computer Engineering, Mokpo National University, South Korea. \\
$^{4}$National Institute of Agricultural Sciences, South Korea.
}
\def\corrAuthor{Corresponding Author}
\def\corrEmail{Jucheng Yang: jcyang@tust.edu.cn, Sook Yoon: syoon@mokpo.ac.kr.}

\begin{document}
\onecolumn
\firstpage{1}

% \title {Challenges in plant disease recognition Using Deep Learning from Real Scenario Perspective}
% \title{Embracing Limited and imperfect dataset: A Review on plant disease recognition Using Deep Learning}
\title{Embrace Limited and Imperfect Training Datasets: Opportunities and Challenges in Plant Disease Recognition Using Deep Learning}

\author[\firstAuthorLast ]{\Authors} %This field will be automatically populated
\address{} %This field will be automatically populated
\correspondance{} %This field will be automatically populated

\extraAuth{}% If there are more than 1 corresponding author, comment this line and uncomment the next one.
%\extraAuth{corresponding Author2 \\ Laboratory X2, Institute X2, Department X2, Organization X2, Street X2, City X2 , State XX2 (only USA, Canada and Australia), Zip Code2, X2 Country X2, email2@uni2.edu}

\maketitle

% \linenumbers

\begin{abstract}
Recent advancements in deep learning have brought significant improvements to plant disease recognition. However, achieving satisfactory performance often requires high-quality training datasets, which are challenging and expensive to collect. Consequently, the practical application of current deep learning-based methods in real-world scenarios is hindered by the scarcity of high-quality datasets. In this paper, we argue that embracing poor datasets is viable and aim to explicitly define the challenges associated with using these datasets. To delve into this topic, we analyze the characteristics of high-quality datasets, namely large-scale images and desired annotation, and contrast them with the \emph{limited} and \emph{imperfect} nature of poor datasets. Challenges arise when the training datasets deviate from these characteristics. To provide a comprehensive understanding, we propose a novel and informative taxonomy that categorizes these challenges. Furthermore, we offer a brief overview of existing studies and approaches that address these challenges. We believe that our paper sheds light on the importance of embracing poor datasets, enhances the understanding of the associated challenges, and contributes to the ambitious objective of deploying deep learning in real-world applications. To facilitate the progress, we finally describe several outstanding questions and point out potential future directions. Although our primary focus is on plant disease recognition, we emphasize that the principles of embracing and analyzing poor datasets are applicable to a wider range of domains, including agriculture.
% A significant finding of our research is the lack of a comprehensive set of desired datasets to evaluate trained models. 

\tiny
\keyFont{ \section{Keywords:} plant disease recognition, AI in agriculture, Deep learning in agriculture, smart agriculture, precision agriculture.}
\end{abstract}

\section{Introduction}

Plant diseases are responsible for significant yield losses \citep{savary2019global}, making their recognition a crucial task in crop cultivation. In the past decade, deep learning, characterized by two essential attributes inherited from classical machine learning methods \citep{kawasaki2015basic, mohanty2016using, fuentes2017robust}, has emerged as a promising approach for this purpose. Firstly, deep learning possesses the remarkable ability to serve as a feature extractor \citep{singh2018deep, bengio2021deep}. This stands in contrast to traditional machine learning, which often necessitates human experts to manually design features, such as histograms of oriented gradients for RGB images \citep{fan2022leaf} and vegetation indices for hyperspectral and multispectral images \citep{abdulridha2020detecting}. However, designing effective features has proven challenging and often requires diversity for different tasks. Secondly, deep learning-based methods have demonstrated "decent performance" in numerous studies on plant disease recognition \citep{singh2018deep, boulent2019convolutional, abade2021plant, liu2021plant, ouhami2021computer, singh2021challenges, thakur2022trends}. Furthermore, the implementation of deep learning on farms offers the enticing advantage of liberating human labor and significantly reducing associated costs. This is particularly valuable in the present century, as the global population is expected to continue increasing, while the number of agricultural workers has been steadily declining.

While deep learning has demonstrated its potential, the requirement for high-quality datasets to achieve satisfactory performance remains a challenge. Unfortunately, collecting such datasets is often prohibitively expensive and extremely challenging in many real-world applications \citep{xu2022style, xu2023comprehensive}. Conversely, poor datasets are prevalent, and current models may struggle when confronted with them. Recognizing this reality, we contend that embracing poor datasets presents new opportunities to advance plant disease recognition in real-world applications. To further enrich the relevant understanding of this embrace, we analyze the characteristics of the desired high-quality datasets: large-scale and desired annotation. Specifically, large-scale datasets provide a vast quantity of information within the images, while desired annotation ensures that the images are annotated in accordance with specific criteria and objectives. More details are discussed in Section \ref{section:analysis}. In contrast, poor-quality datasets are defined by their deviations from the characteristics. Specifically, a dataset not on a large scale is categorized as limited, while a dataset lacking the desired annotation is considered imperfect. Embracing poor datasets, therefore, entails embracing limited and imperfect dataset, each of which is further explored and analyzed in Sections \ref{section:limited} and \ref{section:imperfect}, respectively. The challenges associated with embracing limited and imperfect datasets are explicitly defined within these sections. A novel taxonomy detailing these challenges is conceptually described in Section \ref{section:formluation}, and Table \ref{table:challenges} provides a glimpse of the taxonomy.

\begin{table}[h!]
    \centering
    \small
    \begin{tabularx}{\textwidth}{|l|l|l|X|}
        \hline
        \multicolumn{3}{|c|}{Challenge} & Definition \\
        \hline
        \multirow{4}{*}{ Limited dataset} & \multirow{2}{*}{Class-level} & Few-shot &  All classes have similar few annotated images, where trained models may get low performance for all classes. \\
        \cline{3-4}
        & & Class imbalance & One class has many more annotated images than another class, where trained models may get high performance in the former class but suffer in the latter class. \\
        \cline{2-4}
        & \multirow{2}{*}{Dataset-level} & Domain shift & The training and test datasets share the same label spaces but are in different distribution spaces, where trained models may get low test performance. \\
        \cline{3-4}
        & & Unknown class & Unknown (new) classes exist in the test dataset, where trained models will consider the corresponding image into a known class and not distinguish the unknown from known classes. \\
        \hline
        \multirow{3}{*}{Imperfect dataset}&  \multicolumn{2}{l|}{Incomplete annotation} & Training datasets have labeled and unlabeled images simultaneously, where utilizing the unlabeled images may contribute to the test performance. \\
        \cline{2-4}
        & \multicolumn{2}{l|}{Inexact annotation} & Training datasets are given with only coarse-grained annotations, where utilizing these annotations is challenging to train models. \\
        \cline{2-4}
        & \multicolumn{2}{l|}{Inaccurate annotation} & Some annotations may be inaccurate, where it is challenging to get decent test performance by utilizing these annotations to train models. \\
        \hline
    \end{tabularx}
    \caption{Taxonomy of challenges arising when embracing limited and imperfect dataset. A limited dataset suggests that the training dataset is not on a large scale, while an imperfect dataset implies that the annotations are not expected. Additionally, the challenges of limited dataset encompass class-level challenges, which involve image variations among different classes within the training dataset, and dataset-level challenges, which pertain to the information gap between the training and test datasets.}
    \label{table:challenges}
\end{table}

This study distinguishes itself from existing survey papers on plant disease recognition using deep learning by adopting a "challenge-oriented" approach instead of a "technique-oriented" one. While previous works such as \citep{singh2018deep, boulent2019convolutional, abade2021plant, liu2021plant, ouhami2021computer, singh2021challenges, thakur2022trends} have focused on summarizing existing techniques and relevant materials, we have identified the key challenges associated within this field. Specifically, we highlight the scarcity of large-scale annotated data and advocate for embracing the concept of limited and imperfect datasets when deploying deep learning in real-world applications. 
% Additionally, we emphasize the interconnectedness between general computer vision and plant science. Plant science shares similarities with general computer vision in terms of problem formulations, techniques, and individual characteristics. However, it is essential to consider the unique aspects of plant science and approach them from diverse perspectives. 

To conclude, in pursuit of deploying deep learning for plant disease recognition in real-world applications with satisfactory performance, we offer a perspective that embraces limited and imperfect datasets, contrasting with high-quality data. Our main contributions are as follows:
\begin{itemize}
    \item We \emph{explicitly} argue embracing limited and imperfect datasets for plant disease recognition using deep learning, motivated by the reality that collecting high-quality datasets is expensive and challenging.
    \item We analyze the underlying reasons behind the current necessity for high-quality datasets in Section \ref{section:analysis}.
    \item We present a \emph{taxonomy} of challenges associated with the embrace in Section \ref{section:formluation}, with \emph{formal definitions}. A concise overview of existing studies that tackle them is also given as discussed in Sections \ref{section:limited} and \ref{section:imperfect}.
    \item We provide noteworthy questions and highlight potential directions for further exploration in Section \ref{section:conclusion}.
\end{itemize}

\section{Why Is High-Quality Dataset Desired?}
\label{section:analysis}

In general, to achieve promising performance using deep learning models, the training datasets should have two characteristics: large-scale and annotated with desired strategies. This section aims to probe the underlying reasons behind the desired characteristics.

\subsection{Large-scale Dataset}
\textbf{Deep Learning Models Require Large-scale Data}. For two reasons, deep learning generally requires a large-scale training dataset to obtain a comparable test performance. First, there are enormous learnable parameters in deep learning-based models that require large-scale data \citep{krizhevsky2017imagenet, sarker2021deep}. This ensures that a better feature extractor could be learned; otherwise, the training data points could be remembered, resulting in a poor test performance \citep{xu2023comprehensive}. Second, the distribution of the training dataset is gradually approaching that of the test dataset when the training dataset becomes larger, supporting a better test performance. For example, a model trained with images captured in laboratories is not expected to be effective when tested with images captured on farms \citep{guth2023lab, wu2023laboratory}.

\begin{figure}[h!]
    \centering
    \includegraphics[width=0.7\textwidth]{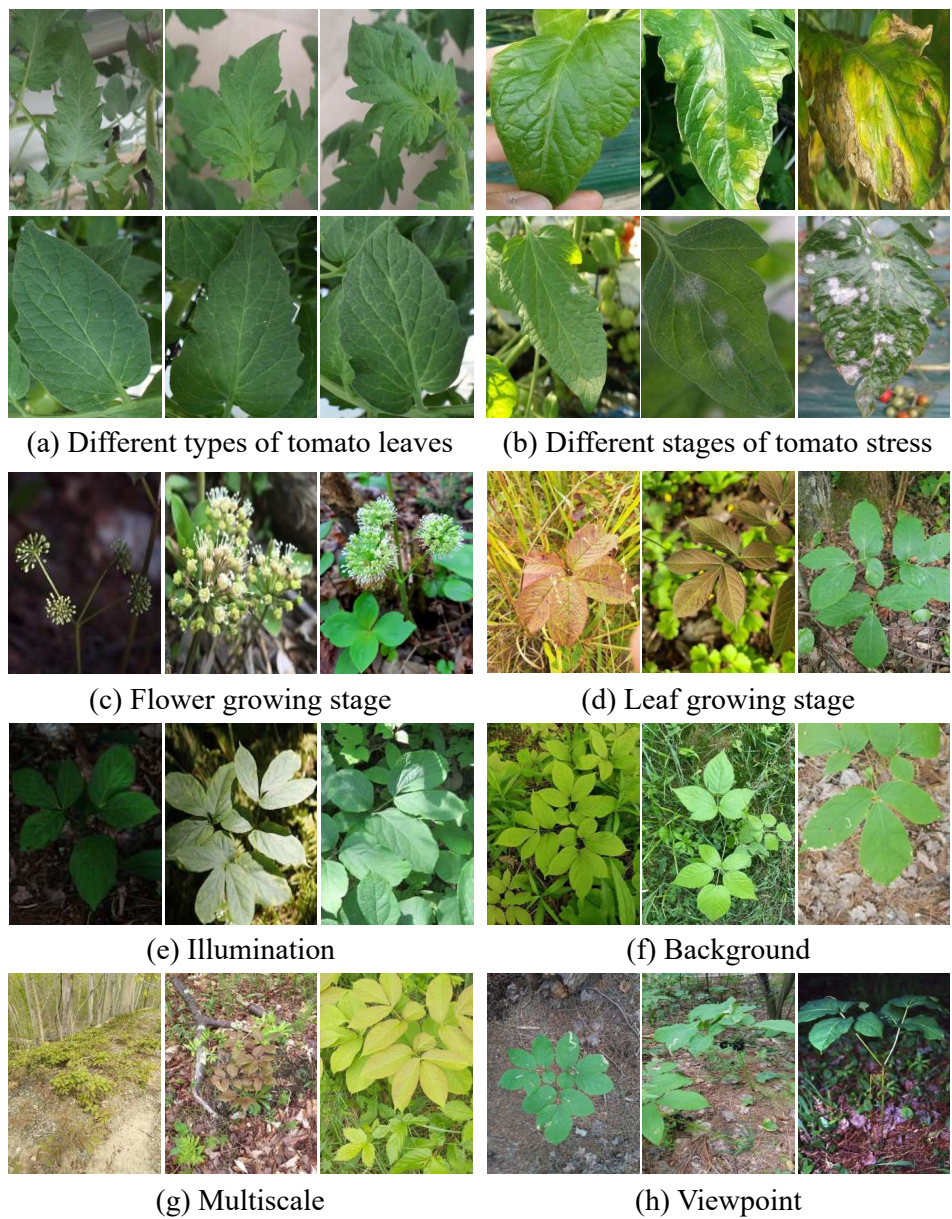}
    \caption{Illustration of the intra-class image variation. The images in (a) and (b) are tomato leaves. Other images stem from a species, Aralia nudicaulis, in the PlantCLEF2022 dataset \citep{goeau2022overview}. Every group suggests that the pictures belonging to the same plant disease may have visual diversities.}
    \label{fig:intra-class}
\end{figure}

\textbf{Huge Image Variation Requires Large-Scale Datasets}. The requirement for a large-scale training dataset comes from not only deep learning but also the task of plant disease recognition, called image variation \citep{xu2023enhanced, xu2023comprehensive}. Considering the types of plant diseases, it can be divided into intra-class, differences within the same plant disease, and inter-class, heterogeneity between the two plant diseases. The intra-class image variation, partially illustrated in Figure \ref{fig:intra-class}, originates from three main elements. The first one is “plant itself”. For example, some plants may have different types, such as different types of tomatoes, with diverse leaf shapes and sizes, as shown in (a) of Figure \ref{fig:intra-class}. The same type of plant and plant disease may also occur at various stages with individual visual patterns. For example, (b–d) in Figure \ref{fig:intra-class} has shown the different stages of plant disease, flowers, and leaves. Second, the plants may be grown in different “environments”, such as fields and greenhouses. Heterogeneous illuminations and backgrounds in the fields are shown in Figure \ref{fig:intra-class} (e) and (f), respectively. Third, ``imaging processing” is another source of intra-class image variation. Arguably, optical sensors and platforms result in greater diversity, such as RGB  and thermal images, phones, and satellites \citep{oerke2014proximal, mahlein2016plant}. When the optical sensors and platforms are fixed, the distance between the plants and sensors results in a multiscale challenge, as shown in (g) of Figure \ref{fig:intra-class}. In addition, these viewpoints also lead to variations, as shown in (h) of Figure \ref{fig:intra-class}.

\begin{figure}[h!]
    \centering
    \includegraphics[width=0.7\textwidth]{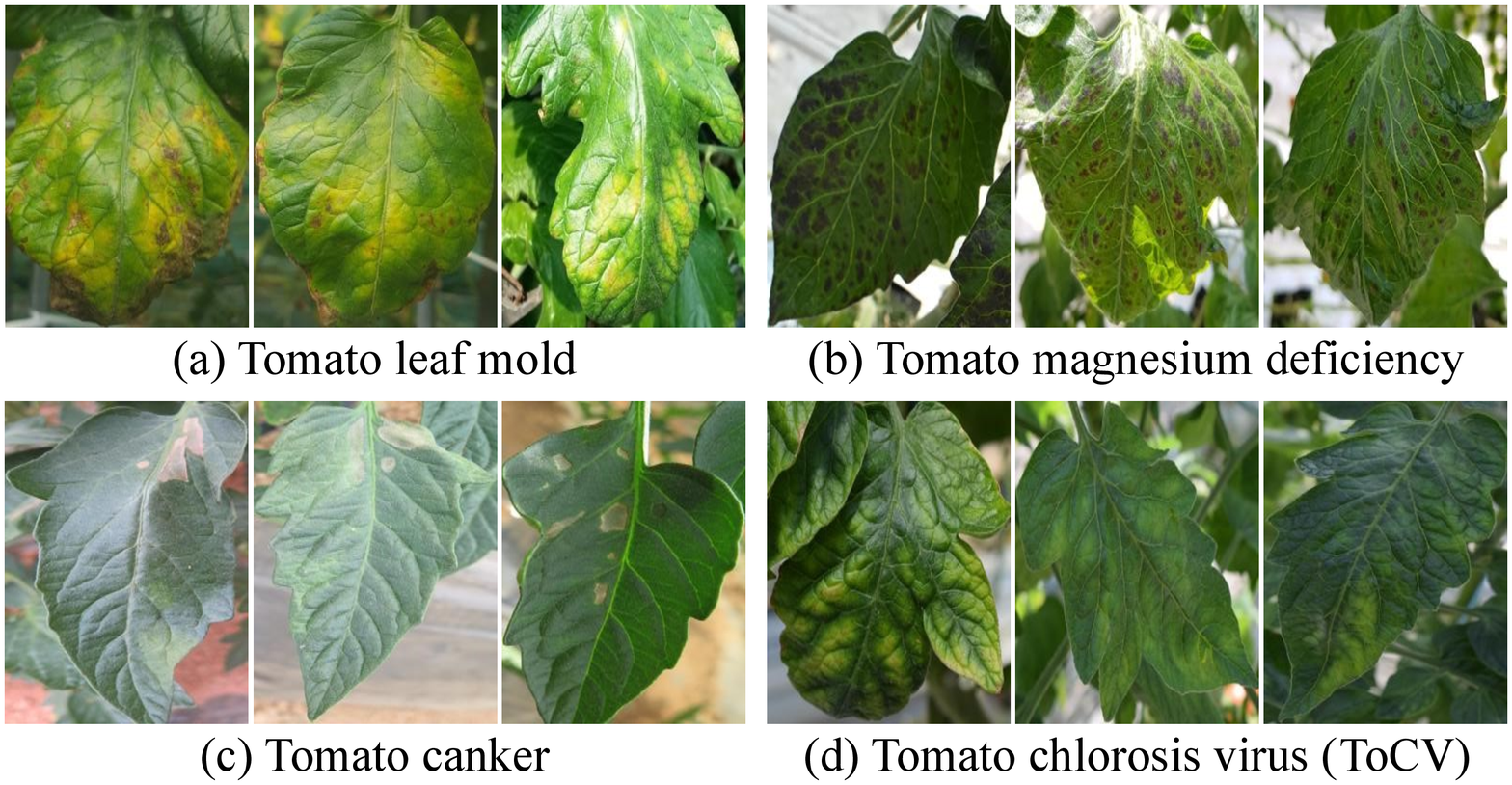}
    \caption{Illustration of inter-class image variation. It can be cast to a relative challenge where the visual deviations between (a) and (c) are larger than that between (b) and (d). A corresponding strategy is to collect more data for (b) and (d) in that models need more evidence to make decisions for hard scenarios.}
    
    \label{fig:inter-class}
\end{figure}

The inter-class variation assumes that one type of plant disease is visually different from another, such as tomato leaf mold and canker, as shown in Figure \ref{fig:inter-class} (a) and (c). We emphasize that this assumption should be considered when formulating the application objectives. For example, early symptoms in RGB images of plant diseases really resemble healthy ones, and consequently, finding related plant diseases at very early stages may not be reliable. In addition, inter-class image variation can be viewed as a relative challenge. Specifically, the visual differences between the two plant diseases could be larger than those of the counterparts of another pair. As shown in Figure \ref{fig:inter-class}, the visual deviations between tomato leaf mold and canker are larger than those between tomato magnesium deficiency and chlorosis virus. A strategy used for this scenario is to collect more data for the close pairs, in which the models required more evidence to make decisions.

\subsection{Desired Annotation Strategy}
\begin{figure}[h!]
    \centering
    \includegraphics[width=0.7\textwidth]{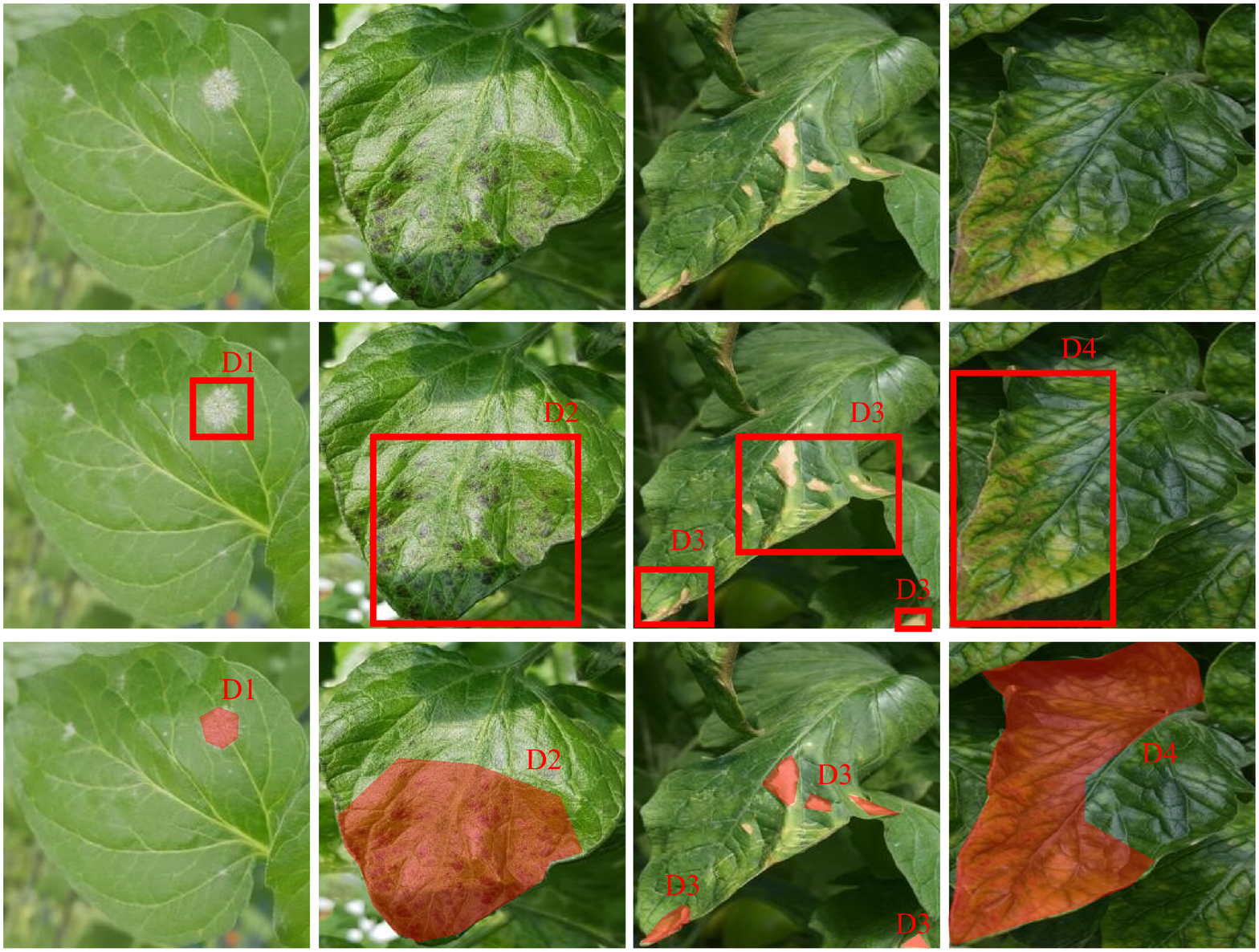}
    \caption{Annotation strategies in three primary tasks of plant disease recognition. From the first to the last rows are image classification, object detection, and segmentation, respectively. $Dm$ $(m=1, 2, 3, 4)$ denote the types of plant disease and every column suggests different plant diseases. In the simplest way, image classification refers to assigning a class to one image whereas, in object detection, classes and their locations (bounding box) are entailed to predict. Segmentation requires class prediction at a pixel level. From the first to the last row, the annotation becomes more complicated and thus more time-consuming. The images in real-world applications tend to be more complex than these examples, such as multiple diseases existing in one leaf and one image including multiple leaves. Further, a desired annotation strategy embraces three primary points: exclusive, extensive, and precise. The exclusive suggests that every annotation just includes one specific visual plant disease pattern whereas the extensive denotes that every plant disease in the images should be annotated. The precise requires that the images should be annotated precisely. Violating the three points leads to the challenges of imperfect annotation.}
    \label{fig:task}
\end{figure}

Deep learning is first trained in a “training dataset” and then tested in a “test dataset”. A “validation dataset” is usually utilized to select the best-trained model from different hyperparameters and other training settings. The training and validation datasets are accessible at both the training and test stages, whereas the test dataset is only accessible at the test stage. Furthermore, the training and validation datasets for most deep learning-based models are hypothesized to be annotated in the desired manner. A desired annotation strategy, called EEP, has three primary points: exclusion, extensiveness, and precision. This has suggested that every annotation included only one specific visual pattern of plant disease, whereas extensive indicated that every plant disease in the images should have been annotated. The last one requires a precise annotation. For example, in image classification, every image should have included a type of plant disease and be linked to a label, as illustrated in the first row of Figure \ref{fig:task}. By contrast, as shown in the second row of Figure 1, object detection generalizes the idea that one image can cover multiple plant diseases. However, every region should be annotated with labels and locations (a bounding box with four values: two for the left point in the horizontal and vertical directions, and two for width and height). Accordingly, segmentation is a task for recognizing plant disease, in which every pixel should be assigned a label, as suggested in the last row of Figure \ref{fig:task}.

\section{Challenge Formulation with Limited and Imperfect Datasets}
\label{section:formluation}
As previously discussed, achieving satisfactory performance using deep learning often requires training datasets that are both annotated as desired and of large scale \citep{lu2022generative}. However, collecting such datasets is frequently challenging, time-consuming, and costly. Consequently, existing models may encounter limitations when applied to real-world scenarios without access to high-quality datasets. Therefore, a more convincing objective is to secure the satisfactory performance of models using limited and imperfect training datasets. However, this concept remains relatively unexplored within the context of plant disease recognition. The present study aims to shed light on this direction, with the ultimate goal of monitoring plant growth, thereby reducing human intervention and potentially mitigating the issues arising from plant diseases.

To comprehend the challenges that arise when dealing with limited and imperfect training datasets, we propose a novel taxonomy. Specifically, the term "limited dataset" refers to scenarios where the training dataset is not on a large scale, while "imperfect dataset" describes situations where the annotations of the training dataset deviate from the expected and desired. The limited dataset can be further divided into two subcategories: class-level, which examines deviations among different classes within the training datasets, and dataset-level, which analyzes the heterogeneity between the training and test datasets. On the other hand, imperfect dataset can be classified into three distinct types based on the nature of conflict: incomplete annotation, where a portion of images lacks annotations; inexact annotation, where some classes are annotated in a coarse-grained manner; and inaccurate annotation, where certain images are annotated with inaccuracies or even incorrect labels. Table \ref{table:challenges} offers a glimpse into this comprehensive taxonomy.

\section{Limited Dataset}
\label{section:limited}
To make the following content easy to understand, several notations are first described. $\mathcal{D}_{X}$ and $\mathcal{D}_{Y}$ denote the training and test datasets, $X$ and $Y$ denoting the training and test domains. Generally, the two datasets encompass $n$ plant disease classes: $c^1, c^2, ..., c^N$. Let $n_{X_i}$ $(i=1, 2, ..., N)$ and $n_{Y_j}$ $(j=1, 2, ..., N)$ denote the numbers of annotated images for class $c^i$ in $\mathcal{D}_{X}$ and class $c^j$ in $\mathcal{D}_{Y}$, respectively.

\subsection{Class-level Limited Dataset}

The class-level limited dataset, confined to the training dataset, is a case in which the number of annotated images for a class is small. Considering the differences across the different classes, they are divided into few-shot and class imbalances.

\subsubsection{Few-shot}

The few-shot challenge assumes that collecting and annotating images are expensive for \emph{every} plant disease with the same number of annotated images. Formally, this challenge is \emph{strictly} defined as: $n_{X_i}=M$, where $M\in \mathbb{R}^+$ is a small natural number. In general, $M$ may be equal to 5 or 10, which suggests that every plant disease has only 5 or 10 annotated images. Moreover, the few-shot challenge could be \emph{generalized} as:
\begin{equation}
    n_{X_i} \approx M,
\end{equation}
where every class contain \emph{approximately} $M$ annotated images. The essential issue is that a few annotated images could not provide sufficient evidence to train a deep learning-based model; thus, the trained model could not be generalized in the test dataset for every class. Based on this, we further extend the few-shot challenge from a small number of annotated images to a larger case, such as 100 and even 500, where \emph{most deep learning-based models could not obtain good test performance for every plant disease}. This motivation is based on the observation that plant disease may have huge intra-class image variation and relatively low inter-class image variation.

To address this few-shot challenge, image manipulation, a set of traditional image processing methods, such as translation and flipping, is one of the simplest methods. It is hypothesized to retain class information and mimic image variations. For example, image translation changes the positions of objects in an image. This method is utilized to increase the number of training images from 350 to 39,010 for six plant diseases and healthy leaves \citep{Gorad_2021}. In particular, a new background is fused to the object of plant disease to create diverse backgrounds in the field rather than in the laboratory \citep{gui2021towards}. Owing to its simple deployment, image manipulation is leveraged by default with many other advanced methods in the general few-shot cases \citep{mohanty2016using, xu2022transfer}. In addition, the image-generating models provide opportunities. Conditional generative adversarial networks (CGAN) \citep{mirza2014conditional} can generate new images, and given a label, the generated images are assumed to be similar to the original images, but not the same \citep{abbas2021tomato}. Intuitively, image-generating models aim to learn image variations in the original training dataset and then create new image variations. However, learning an image-generating model requires data, which is often not satisfactory in the few-shot challenge.

Another effective and efficient method is transfer learning, which transfers knowledge for plant disease recognition from another task with a large-scale training dataset, assuming that learning one task with a large-scale training dataset is beneficial to plant disease recognition \citep{zhuang2020comprehensive}. Generally, the plant disease recognition dataset is called the target dataset and the task is called the target task, whereas another one is termed as source dataset and task, respectively. With this strategy, the first issue is to choose a better model from many deep learning-based models \citep{mohanty2016using, kaya2019analysis, chen2020using}, such as ResNet \citep{he2016deep} and DenseNet \citep{huang2017densely}. The way to adopt a given pre-trained model is a question, such as freezing most of the models and training the remaining part \citep{sethy2020deep}. In addition to models, choosing a better source dataset is another essential issue. A seminal work directly utilized a generic computer vision dataset, ImageNet \citep{deng2009imagenet}. Although it has many variations, it is not that kindly related to plant disease recognition. Simultaneously, a source dataset related to the target dataset is more appealing \citep{matsoukas2022makes, neyshabur2020being}. In this manner, plant-related datasets are considered, such as the AIChallenger2018 \citep{zhao2022identification} and PlantCLEF2017 \citep{kim2021improved} datasets. Furthermore, the image variations inside the source dataset should also be considered. For example, PlantCLEF2017 has more annotated images and higher intra-class variations than AIChallenge2018 where most images are collected in the laboratory. Considering the model and source dataset together is therefore encouraging. Embracing this idea, PlantCLEF2022 \citep{goeau2022overview, xu2022transfer2}, a large-scale plant-relevant dataset with 2,885,052 annotated images for 80,000 classes, is leveraged with a ViT-based \citep{dosovitskiy2020image} model rather than convolution neural network (CNN), to achieve versatile plant disease recognition \citep{xu2022transfer}. With this strategy, a mean test accuracy of 86.29\% over 12 datasets of plant disease recognition in a 20-shot case is achieved with a fast convergence speed, which is 12.76\% higher than the current state-of-the-art accuracy of 73.53\%.

Although transfer learning has been widely adopted, we argue that more opportunities still exist. First, transfer learning involves more segments beyond the models and source datasets, such as the loss function to pre-train and re-train the model. Considering these segments may provide motivation for improving the performance of plant disease recognition. We argue that the new methods in the general field of computer vision are useful for plant disease recognition. For example, a model could be trained using a source dataset with not only annotated images but also paired text to obtain improved semantic representations \citep{radford2021learning, Wei2022MVP}. The utilization of transfer learning in practical applications is another issue. For example, current deep learning-based models have numerous parameters and thus should be compressed for embedding systems. In this case, transferring knowledge from a large to a small model is desirable \citep{abbasi2020compress}.

Other possible strategies, such as meta-learning \citep{huisman2021survey} and metric learning \citep{kaya2019deep}, have received attention over the last few years. Generic deep learning directly outputs the final results such as the type of plant disease, and could be used to learn the relationship between samples and the corresponding ground truth. In contrast, meta-learning aims to learn how to learn, with which the output is rather parameters used to train a new task \citep{chen2021meta, li2021meta, nuthalapati2021multi}. One of the primary issues is that current plant disease recognition datasets could not support an enormous number of source tasks, such as more than 110,000 tasks \citep{chen2021meta}. Furthermore, metric learning attempts to learn the differences between samples. In general, it pushes the samples closer within the same class and away from different classes \citep{li2020few, afifi2020convolutional, egusquiza2022analysis}. Finally, the aforementioned strategies can be combined to further improve the performance, such as \citep{afifi2020convolutional} utilizing transfer learning and metric learning simultaneously.

\subsubsection{Class Imbalance}

The few-shot challenge assumes that each type of plant disease occurs with a similar frequency, thus resulting in small image variations. However, some plant diseases occur at a higher frequency than others in the natural world. For instance, some plant diseases may appear more often than others, and even for one specific type, different stages can be observed with diverse frequencies. In this case, one class may have a much higher number of annotated images than another class in the training dataset, termed a "class imbalance". Mathematically, the class imbalance challenge is formalized as $n_{X_i} \gg n_{X_j}$, where $\gg$ denotes much larger. A class with many more annotated images refers to the majority class; otherwise, it refers to the minority class \citep{xu2022style}. The fundamental challenge is that the trained model tends to assign a high probability to the majority class during the test stage because it contributes more at the training stage \citep{xu2023comprehensive}. However, when the minority class also has many annotated images, the models may exhibit acceptable performance. Therefore, we propose a strict definition that is closer to the real applications:
\begin{equation}
\left\{
    \begin{array}{l}
      n_{X_i} \gg n_{X_j},\\
      n_{X_j} \leq M.
    \end{array}
  \right.
\end{equation}

In the strict version, the number of annotated images of the minority class is not only much less than that of the majority class but should also be lower than a specific value. We argue that $M$ should not be fixed for all tasks. By contrast, this value depends on multiple factors, such as intra- and inter-class variations. Essentially, \emph{deep learning-based models may not be able to learn robust features for the minority class in the class-imbalance datasets}. To mitigate this challenge, the primary idea is to increase the performance of the minority class while maintaining that of the majority class.

Theoretically, the majority of strategies designed for few-shot could be utilized in that class imbalance becomes a few-shot challenge when the number of annotated images of the majority class reduces to a certain extent. This subsection introduces the methods aiming to specifically facilitate the minority classes. Compared to the majority class, the lower performance of the minority class is assumed to be due to the lower observation frequency of the models. This assumption inspires balancing the frequency for models by pushing the model to look at the images of the minority class more often \citep{nafi2020addressing}. Similarly, models can also be punished more by using samples from the minority class \citep{nafi2020addressing, oksuz2020imbalance}. In addition, image augmentation aims to increase image variations to facilitate deep learning-based models. The methods belonging to image augmentation for the class imbalance differing from that of the few-shot is the basic motivation that the majority class can be leveraged for the minority. Conditional image-generating models implicitly utilize this insight by training a single model to learn from all classes \citep{mirza2014conditional, abbas2021tomato}. By contrast, translating an image from one class to another directly utilizes the information among these classes \citep{nazki2020unsupervised, cap2020leafgan, lu2022generative}. To further consider the intra-class image variation from the majority class to the minority class, a specific loss is leveraged along with the image translation strategy \citep{xu2022style}.

\subsection{Dataset-level Limited Dataset}

The limited dataset at the class level considers situations among the classes of the training dataset, whereas heterogeneity between the training and test datasets appears at the dataset level. It is further categorized into unknown classes and domain shifts. The former suggests that some classes in the test dataset, termed unknown classes, do not appear in the training dataset; whereas the latter emphasizes that the image variations in the test and training datasets are diverse. We emphasize that these two categories focus on specific essences and that their combination may exist at a higher frequency in real-world applications.

\subsubsection{Unknown Class}

The class in the training dataset refers to the "known class" while a class existing in the test dataset but not in the training dataset is termed an "unknown class" \citep{geng2020recent}. In the concept of plant disease, unknown classes may result in a large economic loss, and recognizing them is thus one of the fundamental demands. Simultaneously, collecting all the existing plant diseases is difficult and even impossible for real-world applications. Therefore, assuming the existence of unknown classes in the test dataset is encouraging. In this scenario, the task of plant disease recognition has two-fold; classifying known classes, and rejecting unknown classes \citep{yang2021generalized}. This task refers to open set recognition or out-of-distribution and has witnessed significant developments in the field of computer vision \citep{geng2020recent, yang2021generalized, salehi2021unified}. However, it has been rather underdeveloped for recognizing plant disease. In the following paragraphs, three key understandings from the computer vision field are first introduced, followed by a review of the literature on plant disease recognition.

First, thresholds are commonly employed to distinguish unknown classes, such as when an image larger than the threshold is categorized as known. In this case, two things are essential: a method to compute a value for a given image and a method to set a threshold. Currently, the probability \citep{liang2018enhancing}, energy \citep{zhang2020hybrid}, and reconstruction error \citep{sun2020conditional} are the three main strategies for a given image. For example, known classes are assumed to have higher probabilities, lower energies, and smaller reconstruction errors than unknown classes. Simultaneously, a fixed threshold tuned in the training dataset is widely employed, such as the accuracy to maintain 95\% of the images in the training dataset as known \citep{huang2022class}. This fixed one can be deemed at the dataset level and the class-level threshold has been recently considered in that different known classes probably behave diversely \citep{wang2022vim}.

Secondly, learning a good classifier with known classes is an effective and efficient strategy, such as utilizing strong image augmentation and longer training times \citep{vaze2022openset}. A good classifier requires models to learn a robust feature space to distinguish one known class from another. Generally, a robust feature space is tight for a specific class and the distances between the two classes are sufficiently large, with which unknown classes have more possibilities to be recognized. However, known classes with occlusions and unknown classes with features similar to those of known classes trigger problems in this strategy \citep{dietterich2022familiarity}. Finally, the exposure of potential novel classes, not the unknown class in the test dataset, is a convincing strategy because the primary challenge is that models are trained with only known classes and extra information about unknown classes can provide extra information \citep{dietterich2022familiarity, zhou2021learning}. With this paradigm, the potential unknown classes and efficiency of sampling images from unknown classes are essential \citep{chen2021adversarial}.

In plant disease recognition, an existing work aims to learn a good classifier via metric learning \citep{you2022deep}, with the inspiration that the distances between two images from the same unknown class should be smaller than those from different known classes. Generally, metric learning pushes models to learn robust feature spaces and thus implicitly contributes to the recognition of the unknown class. In addition, an extra probability branch is explicitly utilized to distinguish between known and unknown classes along with a generic classification branch for known classes \citep{jiang2022detection}. Simultaneously, images belonging to unknown classes are utilized to train the models, where exposure to unknown classes is beneficial, although unknown classes in the training stage may also appear in the test stage. The ratio between the number of known and unknown classes is formally analyzed \citep{fuentes2021open}. The experimental results suggest that the performance deteriorates with more unknown classes mainly because of the shortage of useful information in the training dataset.

\subsubsection{Domain Shift}

Domain shift is a common problem in deep learning where the training and test data come from different domains. In such cases, the trained model may perform poorly on the test data, resulting in a phenomenon known as a "domain shift". Domain shifts can occur for several reasons such as differences in data distribution, scale, and quality. The general assumptions are as follows:
\begin{equation}
    P(c|X) \neq P(c|Y),
\end{equation}
where $P(c|X)$ represents the probability distribution of one plant disease $c$ given the input images in the training dataset $X$, and $P(c|Y)$ represents the probability distribution of $c$ given the input images in test dataset $Y$. The inequality sign indicates that the two distributions are not equal, implying that the domain shift can lead to a significant decrease in the performance of the test data, making the model ineffective. The unknown class in the test dataset is a special form of the domain shift challenge but in this section, we aim to highlight the domain shift where the set of classes in the test dataset is a proper subset of that in the training dataset.

In the case of plant disease data, symptoms do not have well-defined boundaries, and gradually change from healthy normal conditions to diseased regions, making it difficult to create homogeneity in the data \citep{barbedo2018factors}. In addition, the inter- and intra-class image variations, as well as the explicit variations given by the domain used for the data collection, add complexity to the model. A frequent performance drop occurs when a model is trained on a dataset from a particular scenario but is further tested on data from a different scenario. A common scenario, for example, is that the training dataset is collected in one place by one person and the test dataset is collected in another place with different infrastructures and illumination by another person with their individual habit of taking pictures, such as training in the images collected in the laboratory and testing in the real-world \citep{guth2023lab, wu2023laboratory}.

To address the domain shift problem, researchers have developed several techniques, such as domain adaptation \citep{wang2018deep} and domain generalization \citep{wang2022generalizing}. Domain adaptation aims to adapt a model to the test domain by modifying the training data or the model itself, whereas domain generalization aims to train a model that can perform well in unseen domains. The goal of generalization is to design a model that can operate efficiently in the same environment or across multiple environments. There are several approaches to domain adaptation for plant disease recognition using deep learning. One approach is to use transfer learning, which involves fine-tuning a pre-trained model on a new dataset. This approach can be effective when the new dataset is similar to the original dataset but may not work as well when there are significant differences between the domains. Another approach is to use domain adaptation techniques such as adversarial training or domain adaptation networks. These techniques involve training a model to recognize features specific to the target domain while also minimizing the differences between the source and target domains. This approach can be effective when there are significant differences between domains but may require more computational resources and training data.

In the literature, this problem has barely been investigated in plant disease recognition; however, it is an important issue for developing a more generalized model. Early work in this area \citep{fuentes2021improving} shows the benefits of using control classes, such as background and healthy leaves, to lead the learning process towards classes of interest. It exhibited improved performance as an easier-to-adapt model across environments. However, data from different backgrounds and environments are required to achieve this goal. This issue was further investigated by \citep{fuentes2021open}, where a bounding box detector was trained to obtain the regions of interest. Then in the second stage, a domain adaption model obtained the features of data from a source farm with known diseases and transferred them to a target farm in which unlabeled data was used to assess the generalization capabilities of the model to recognize regions belonging to known classes or otherwise assigned them as unknown. This scenario showed how implementation could improve the recognition of target diseases and precisely estimate novel information by associating them with an unknown class.

Another important assumption is to address domain shifts across crops and environments. Shibuya et al. \citep{shibuya2021validation} utilized more than 221,000 labeled leaf images from different regions and crops to investigate the performance bias of evaluation within the same farm and the effect of ROI detection. They found that even with many training images, the diagnostic performance for images in fields different from the training images is greatly degraded owing to covariate shifts. From this study, two important questions arise: first, what is the importance of data taken in different environments than the training data for evaluation, and second, how do the pre-detection of regions of interest, including symptoms of diseases, affect the performance? Another essential aspect to investigate in the domain shift is the changes that occur in data collected in the laboratory compared with data collected under field conditions. The generalization capabilities of CNNs are investigated to learn the clear patterns from lab conditions which are to be detected again under new and more complex field conditions while avoiding overfitting \citep{guth2023lab}. The important insight derived from this is that in order to create useful tools for disease detection and classification using deep learning for image analysis, it is crucial to develop a final product that can handle a wide range of images from various crop conditions and locations, including inter-class and intra-class variations \citep{wu2023laboratory}. This requires carefully designed datasets with a large number of image samples that can accommodate the significant variability in crop conditions in different areas.

In summary, domain shift is a critical challenge in plant disease recognition using deep learning. Developing models that can adapt to different domains is essential for building robust and accurate systems that can be used across a wide range of crops and regions. Although there are several approaches to the domain shift, researchers must continue to develop new techniques and datasets for ensuring that these models are effective in real-world scenarios.

\section{Imperfect Dataset}
\label{section:imperfect}
\begin{figure}[h!]
    \centering
    \includegraphics[width=0.7\textwidth]{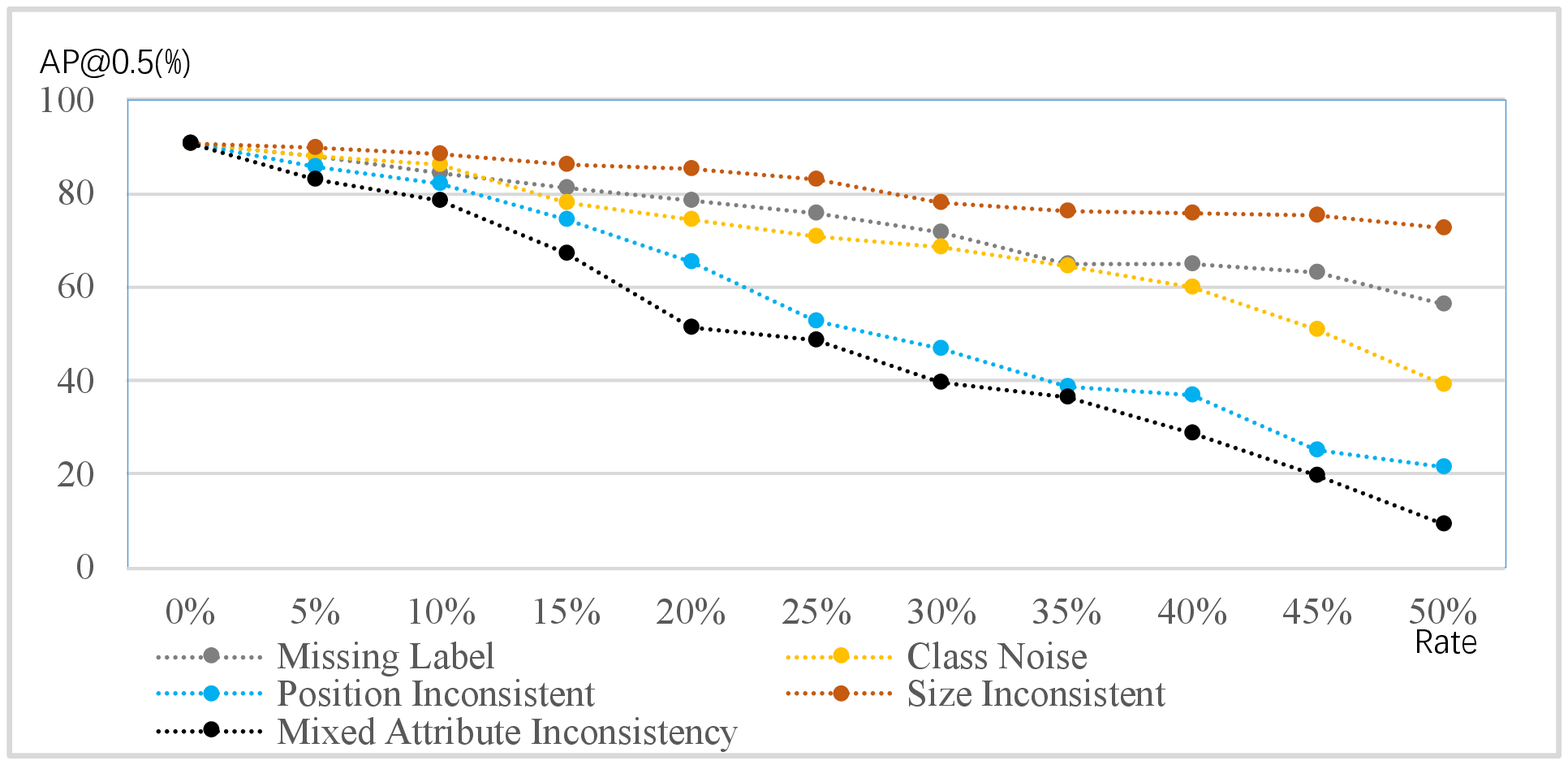}
    \caption{Performance comparison of object detection in a plant disease dataset, using different annotations, this figure adapted from \citep{dong2022data}. The cases of missing labels and class noise suggest some patterns of plant disease have no labels and wrong labels. The inconsistencies of position and size suggest that the position and size are different from the desired. The mixed case is the combination of previous cases. Detection performance clearly degrades when the deviations from the desired ones are severe.}
    \label{fig:annotation}
\end{figure}

The limited dataset challenge considers annotated images within either a class or dataset in which all images in the training dataset are assumed to be annotated, whereas the imperfect dataset challenge instead hypothesizes that the annotation in the training dataset can be missing and not perfect. The imperfect dataset challenge is categorized into incomplete, inexact, and inaccurate annotations based on the violation of the EEP annotation strategy. Figure \ref{fig:annotation} provides a quick impression of the changes in performance when the annotations are different from the desired ones.

\subsection{Incomplete Annotation}

Incomplete annotation indicates that the training dataset includes annotated and unannotated images, primarily because of economic issues and the annotation requirements of expert knowledge in plant science. The number of annotated images is much lower than that of unannotated images. A straightforward method is to discard unannotated images and just use the annotated images to train the models. By contrast, the use of unannotated images has become an active topic in recent years. One strategy to use unlabeled images is self-supervised learning (SSL) which aims to learn better representations, followed by a fine-tuning process within the annotated images. In SSL, a pretext task should first be defined without using annotation to train a deep learning-based model \citep{jing2020self}. Currently, there are many types of pretext tasks, but only a few are utilized to recognize plant disease. In particular, image augmentation does not change one image's type of plant disease and is deemed as a pretext task \citep{nagasubramanian2022plant}. Further, advanced image augmentation methods, such as Mixup \citep{zhang2018mixup} changing the labels linearly, can also be utilized as a pretext task \citep{monowar2022self} using the connections before and after image augmentation.

Semi-supervised learning, which is another active topic, attempts to leverage unlabeled and labeled images simultaneously. One branch directly adopts SSL methods along with a supervised loss function such as softmax. Another branch generates pseudo-labels for unlabeled images, where the labeled images can be leveraged to annotate the unlabeled images by first training a classifier \citep{li2021semi}. In contrast, clustering methods find the similarity without label information and link the images in the test dataset to those in the training dataset \citep{fang2021self}.

Furthermore, active learning aims to select informative images labeled by humans later instead of machines, hoping to annotate fewer images yet obtain a better performance \citep{ren2021survey}. Therefore, selecting informative images effectively and efficiently \citep{nagasubramanian2021useful} is essential. Moreover, the involvement of human experts in the training loop is difficult and inconvenient in real-world applications.

\subsection{Inaccurate Annotation}

Inaccurate annotation, also called noisy annotation, denotes that some annotations in the training dataset may not be correct \citep{algan2021image, dong2022data}. For example, plant disease is incorrectly annotated in the classification case, considering that experts may have conflicting decisions for a given image. Similarly, the bounding box used for object detection may be imprecise. Inaccurate annotation can be mitigated using multiple annotators, but this is expensive. Existing work \citep{dong2022data} suggests that inaccurate annotation results in worse recognition performance, and different noise magnitudes have diverse impacts. Accordingly, facilitating the training process to reduce its impact is a straightforward approach \citep{li2019learning}. Following this idea, new plant diseases are randomly generated for every image, and meta-learning is adopted to obtain consistent predictions \citep{zhai2022rectified}. In this manner, meta-learning aims to reduce the adverse impact of randomly generated labels. Although inaccurate annotations are facilitated, the corresponding images do not contribute to the trained models. Therefore, we highlight inaccurate annotations and employ relevant images by re-correcting the annotations \citep{wang2022narrowing, liu2022robust}, although this idea has not yet been leveraged in plant disease recognition.

\subsection{Inexact Annotation} 
Inexact annotation refers to coarse-grained annotation and the meaning varies for different tasks \citep{zhou2018brief, zhang2021weakly}. For example, only image-level labels are accessible for object detection and segmentation, without bounding boxes re pixel-level classes respectively. We emphasize that image classification also has a situation of inexact annotation, such as multiple diseases existing in one image but only one disease label. Simultaneously, inexact annotations may appear along with precise annotations. A basic assumption in using inexact annotation is that a deep learning-based model may learn a significant area with coarse-grained annotations. Specifically, the activation value in every layer indicates that the pixels contribute to the final prediction diversely. Therefore, computing the most important pixels in an input image is one way of determining the exact annotation. This strategy has been employed for object detection of crop pests \citep{bollis2022weakly} and segmentation of foliar diseases \citep{yi2021lesion}. However, this challenge currently receives less attention than incomplete and inaccurate annotations.

\section{Concluding Remarks and Future Perspectives}
\label{section:conclusion}
In this study, we advocated for embracing limited and imperfect training datasets for plant disease recognition using deep learning, acknowledging the practical difficulties, expenses, and challenges associated with collecting high-quality datasets in real-world applications. While this embrace is more convincing and practical, it also introduces new challenges. To enrich our understanding, we proposed a novel taxonomy of challenges with formal definitions. Additionally, we provided a concise overview of strategies to address these challenges. One noteworthy finding is the limited research focused on dataset-level challenges related to limited datasets and imperfect annotation, while significant developments have been made concerning class-level limited datasets. Another essential point we discovered is the severe shortage of benchmark datasets specifically tailored for real-world applications. By highlighting these insights, we aim to contribute to the advancement of deep learning techniques in real-world applications and foster progress in the domain of plant disease recognition. While the primary focus of this study was on plant disease recognition, we emphasize that the concept of embracing limited and imperfect datasets is applicable to broader fields, such as deep learning in agriculture.

\begin{figure}[h!]
\begin{center}
\includegraphics[width=0.65\linewidth]{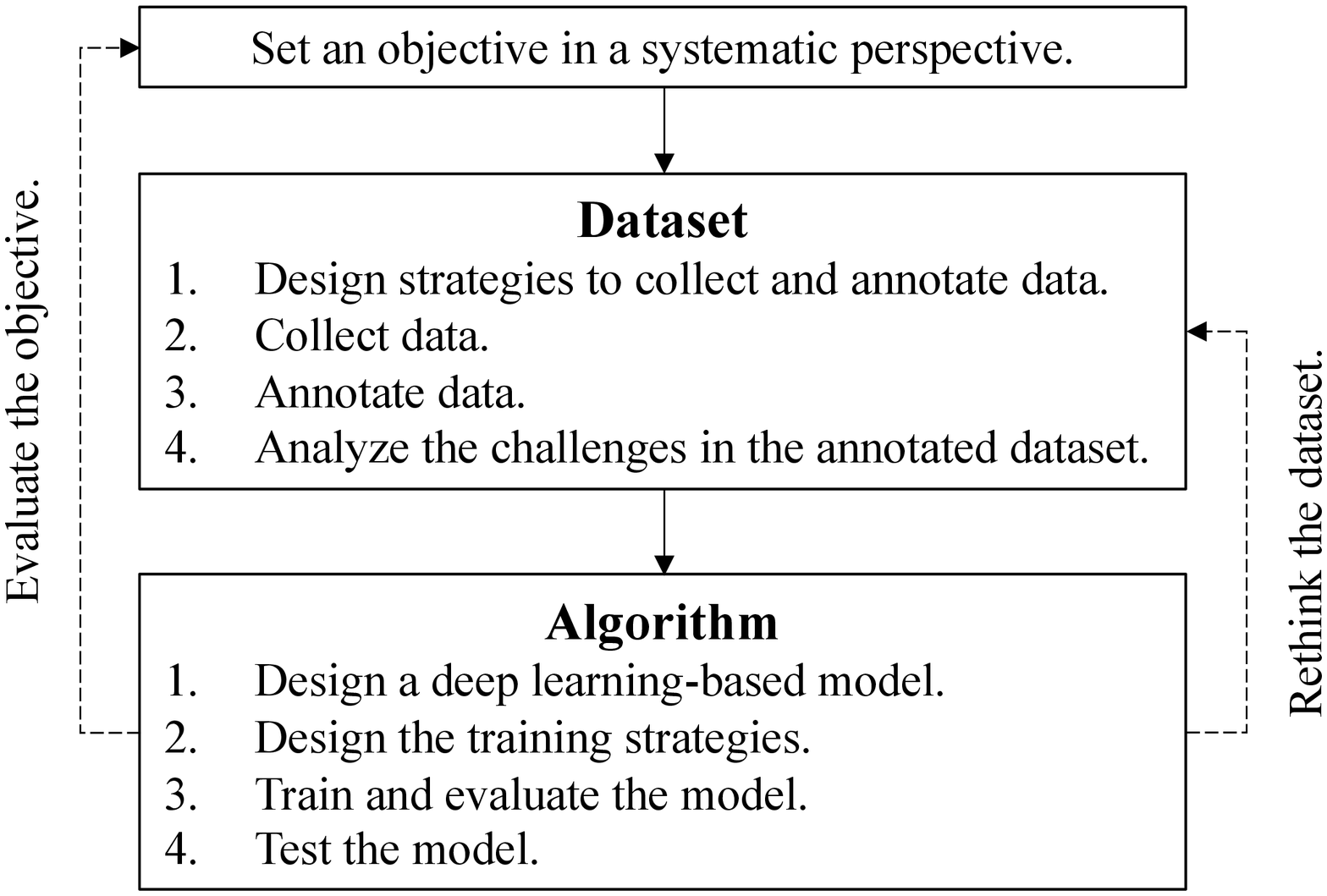}
\end{center}
\caption{Flowchart to deploy deep learning in plant disease recognition. The evaluation of the project objectives and rethinking of the datasets are highlighted.}
\label{fig:flowchart}
\end{figure}

Building upon the challenges posed by limited and imperfect datasets, we propose a process tailored for real-world applications, depicted in Figure \ref{fig:flowchart}. Our objective is to emphasize the importance of evaluating and reevaluating the objectives and datasets. A fundamental assumption is that each class exhibits distinguishable visual patterns in the image space. However, certain classes may share remarkably similar patterns, making them challenging to distinguish. In such cases, the objectives of utilizing deep learning or the collected datasets should be polished, possibly by incorporating novel evidence. We also present several outstanding questions in Box 1 and outline potential future directions in Box 2, seeking to foster further research and advancements in the domain.

\begin{mdframed}[nobreak=true]
Box 1. Outstanding questions.
\begin{itemize}
    \item How can efficiently integrate plant science, including plant disease recognition, and artificial intelligence knowledge from collecting data to deploying a deep learning model?
    \item How to make a reliable dataset for the application-orientated challenges.
    \item Is there any other challenge to deploying deep learning in plant disease recognition, except the limited and imperfect dataset?
    \item What are the heterogeneities between plant disease recognition and generic computer vision tasks?
    \item How to design a preliminary automatic prototype to recognize plant disease as a real-world application?
    \item Considering the success of the large language models and foundation models, what can be done in plant disease recognition and the plant science field?
\end{itemize}
\end{mdframed}

\begin{mdframed}[nobreak=true]
Box 2. Future directions.
    \begin{itemize}
    \item Inspiration
        \begin{itemize}
            \item[$\circ$] Adopt the commonness between plant disease recognition and general computer vision tasks and then adapt the suitable concepts such as new problem formulations and methods.
            \item[$\circ$] Distinguish plant disease recognition from general computer vision tasks such as different plants having similar diseases and further leverage the difference.
        \end{itemize}
    \item Dataset
        \begin{itemize}
            \item[$\circ$] Collect application-orientated datasets, such as for the domain shift.
            \item[$\circ$] Collect datasets from multiple sensors simultaneously, such as RGB and multispectral images.
            \item[$\circ$] Collect datasets from multiple observations such as spatial and temporary, inspired by the effectiveness of accumulated evidence.
            \item[$\circ$] Develop strategies to integrate datasets from the whole community.
        \end{itemize}
    \item Model and algorithm
        \begin{itemize}
            \item[$\circ$] Develop strategies for specific challenges, such as for open set recognition.
            \item[$\circ$] Fine-tune large pre-trained models to achieve better performance in plant disease recognition tasks, and design strategies to achieve parameter-efficient fine-tuning (PEFT).
            \item[$\circ$] Employ small models for the embedding system.
            \item[$\circ$] Integrate large and small models to have decent performance yet for the embedding system.
        \end{itemize}
    \item Analysis
        \begin{itemize}
            \item[$\circ$] Analyze the challenges in a dataset quantitatively.
            \item[$\circ$] Analyze the impacts of strategies to annotate datasets.
            \item[$\circ$] Analyze the relationship between performance, data quality and amount, computing resources, and model capacity. 
        \end{itemize}
    \item Application
        \begin{itemize}
            \item[$\circ$] Evaluate plant disease quantitatively, such as object detection and segmentation.
            \item[$\circ$] Deploy deep learning in real-world applications, such as robotic systems.
            \item[$\circ$] Design versatile plant disease recognition such as multi-plant and multi-dataset, rather than individual models for specific plants and datasets.
            \item[$\circ$] Consider plant disease recognition with other plant-related tasks, such as plant identification.
        \end{itemize}
    \end{itemize}
\end{mdframed}

\section*{Conflict of Interest Statement}
The authors declare that the research was conducted in the absence of any commercial or financial relationships that could be construed as a potential conflict of interest.

\section*{Author Contributions}
Mingle Xu: Conceptualization, Formal analysis, Writing - original draft, Writing - review \& editing.
Hyongsuk Kim: Project administration, Funding acquisition, Writing - review \& editing.
Jucheng Yang: Writing - review \& editing.
Alvaro Fuentes: Writing - original draft, Writing - review \& editing.
Yao Meng: Writing - original draft, Writing - review \& editing.
Sook Yoon: Supervision, Writing - review \& editing.
Taehyun Kim: Resources.
Dong Sun Park: Supervision, Project administration, Funding acquisition, Writing - review \& editing.

\section*{Funding}
This work was partially supported by the Korea Institute of Planning and Evaluation for Technology in Food, Agriculture and Forestry(IPET) and Korea Smart Farm R\&D Foundation(KosFarm) through the Smart Farm Innovation Technology Development Program, funded by the Ministry of Agriculture, Food and Rural Affairs(MAFRA) and Ministry of Science and ICT (MSIT), Rural Development Administration(RDA) (1545027177). This research was partially supported by Basic Science Research Program through the National Research Foundation of Korea(NRF) funded by the Ministry of Education (No. 2019R1A6A1A09031717). 

\section*{Acknowledgments}
We thank Jiuqing Dong (ORCID: 0000-0002-5148-9817) for the early discussions and recommended materials.

% \section*{Supplemental Data}
%  \href{https://www.frontiersin.org/guidelines/author-guidelines#supplementary-material}{Supplementary Material} should be uploaded separately on submission, if there are Supplementary Figures, please include the caption in the same file as the figure. LaTeX Supplementary Material templates can be found in the Frontiers LaTeX folder.

\section*{Data Availability Statement}
This study does not use a dataset.
% Please see the availability of data guidelines for more information, at https://www.frontiersin.org/guidelines/policies-and-publication-ethics#materials-and-data-policies

\bibliographystyle{Frontiers-Harvard} %  Many Frontiers journals use the Harvard referencing system (Author-date), to find the style and resources for the journal you are submitting to: https://zendesk.frontiersin.org/hc/en-us/articles/360017860337-Frontiers-Reference-Styles-by-Journal. For Humanities and Social Sciences articles please include page numbers in the in-text citations 
\bibliography{test}

%%% Make sure to upload the bib file along with the tex file and PDF
%%% Please see the test.bib file for some examples of references

%%% If you don't add the figures in the LaTeX files, please upload them when submitting the article.
%%% Frontiers will add the figures at the end of the provisional pdf automatically
%%% The use of LaTeX coding to draw Diagrams/Figures/Structures should be avoided. They should be external callouts including graphics.

\end{document}